\documentclass[runningheads]{llncs}

\usepackage{eccv}
\usepackage{marvosym}

\usepackage{eccvabbrv}

\usepackage{graphicx}
\usepackage{booktabs}

\usepackage[accsupp]{axessibility}  

\usepackage{hyperref}

\usepackage{orcidlink}

\begin{document}

\title{Edge-Efficient Image Restoration: Transformer Distillation into State-Space Models} 


\author{
Srinivas Soumitri Miriyala$^{1,*,\textsuperscript{\Letter}}$ \and
Sowmya Vajrala$^{1,*,\textsuperscript{\Letter}}$ \and
Sravanth Kodavanti$^{1,*,\textsuperscript{\Letter}}$ \and
Vikram Nelvoy Rajendiran$^{1}$ \and
Sharan Kumar Allur$^{1}$\\
}

\authorrunning{S. Miriyala et al.}

\institute{Samsung Research Institute Bangalore, India}
\maketitle

\setcounter{footnote}{0}
\begingroup{
\let\thefootnote\relax\footnotetext{\textsuperscript{\Letter}{srinivas.soumitri@gmail.com, v.lahari@samsung.com, k.sravanth@samsung.com}} 
\let\thefootnote\relax\footnotetext{\textsuperscript{*}Equal Contribution}}

\begin{abstract}
We propose a modular framework for hybrid image restoration that integrates transformer and state-space model (SSM) blocks with a focus on improving runtime efficiency on edge hardware. While transformers provide strong global modeling through self-attention, their attention kernels incur substantial latency on mobile devices, especially for high-resolution inputs. In contrast, SSMs such as Mamba offer linear-time sequence modeling with lower runtime overhead but may underperform on fine-grained restoration tasks. To balance accuracy and efficiency, we train lightweight SSM blocks as feature-distilled surrogates of transformer blocks and use them to construct hybrid U-Net-style architectures. To automatically discover effective block combinations, we introduce \textbf{Efficient Network Search (ENS)}, a multi-objective search strategy that selects task-specific hybrid configurations from pre-aligned components. ENS optimizes restoration quality while penalizing transformer usage, serving as a lightweight proxy for latency and enabling architecture discovery without repeated hardware profiling. On a Snapdragon~8~Elite CPU, the Restormer baseline requires \textbf{10119.52\,ms} for inference. In contrast, ENS-discovered hybrids significantly reduce runtime: \textbf{ENS-Deblurring} runs in \textbf{2973\,ms} (\textbf{3.4$\times$ faster}), \textbf{ENS-Deraining} in \textbf{5816\,ms} (\textbf{1.74$\times$ faster}), and \textbf{ENS-Denoising} in \textbf{8666\,ms} (\textbf{1.17$\times$ faster}), while maintaining competitive restoration quality.
\end{abstract}

\section{Introduction}
\label{sec:intro}

\begin{figure*}
\centering
\includegraphics[width=\linewidth]{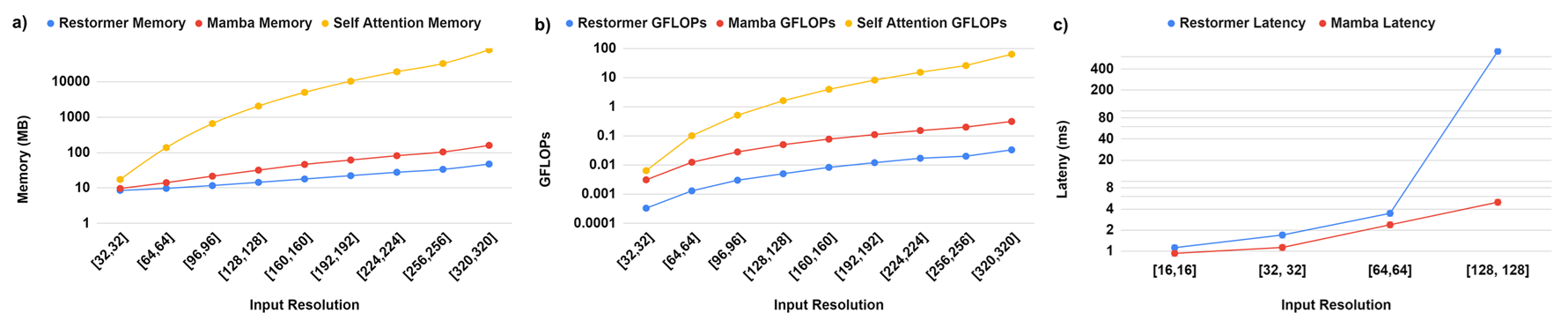}
\caption{a) Computational characteristics of Transformers (quadratic self attention), Mamba, and Restormer (linear Transformer) as profiled on Qualcomm chipset SM8750 on Android smartphone. The Y axis in all three plots is in log-scale. It can be observed that FLOPs and memory consumption are lower for Mamba than Transformer, whereas, Restormer is far more efficient than Mamba. However, when profiled for on-device latency it can be clearly seen that Mamba is the most efficient alternative among the three.} 
\label{fig0}
\end{figure*}

Transformer-based architectures have demonstrated strong performance across a variety of vision tasks, including image restoration, due to their ability to model long-range dependencies via self-attention~\cite{vaswani2017attention}. However, the quadratic complexity of standard attention makes them inefficient for high-resolution inputs, particularly on edge devices. To address this, several linear-scaling transformer variants—such as Linformer~\cite{wang2020linformer}, Restormer~\cite{zamir2022restormer}, and Performer~\cite{choromanski2020rethinking}—reduce attention cost via approximation or architectural simplification. While effective, these designs often encode task-specific inductive biases; for example, Restormer’s gated attention favors spatial patterns, while Linformer assumes low-rank projections suited to language modeling. Such constraints may limit their generalizability.

State-space models (SSMs) like Mamba~\cite{gu2023mamba} offer an alternative, enabling linear-time modeling via implicit memory and recurrent dynamics. Mamba shows promise in long-context tasks~\cite{gu2023combining}, but its application to dense vision remains limited~\cite{yu2024mambaout}, and performance still lags behind transformers on high-fidelity detail restoration~\cite{li2024matir}. Recent adaptations such as MambaIR~\cite{chen2024mambair}, MaIR~\cite{li2024mair} and MatIR~\cite{li2024matir} extend SSMs to pixel-level modeling, leveraging spatial priors and hierarchical encoders. However, in challenging image restoration tasks, these SSMs underperform relative to transformers~\cite{li2024matir}. 

Restormer, a highly efficient linear variant of transformers achieves lower FLOPs and memory usage than both vanilla Transformers and Mamba as can be seen in Figure ~\ref{fig0}, yet its attention mechanism still incurs substantial latency on edge hardware. Our block-level profiling on a commercial mobile chipset shows a consistent trend: as input resolution increases, Mamba blocks exhibit noticeably lower runtime despite their comparable or higher FLOPs. This mismatch reflects the hardware inefficiency of attention-dominated operations relative to Mamba’s linear state-space updates. Such observations motivate the need for hybrid architectures that preserve Transformer accuracy while benefiting from the practical latency advantages of Mamba. While we establish Mamba's low latency, execution-friendly and compute efficient nature, a natural question that arises is whether the representational capacity of a Restormer block can, in practise, be distilled into a Mamba block?

To study the effectiveness of the distillation process, we performed an Effective Reception Field analysis (ERF)~\cite{luo2016understanding} (see Figure ~\ref{fig1}). We observe that while transformers exhibit wide, uniform receptive fields, vanilla SSM blocks remain local. However, after distillation, the SSM ERFs expand significantly, and fine-tuning further narrows the gap. This indicates that global attention behavior can be transferred through feature-level supervision. 
\begin{figure*}
\centering
\includegraphics[width=\linewidth]{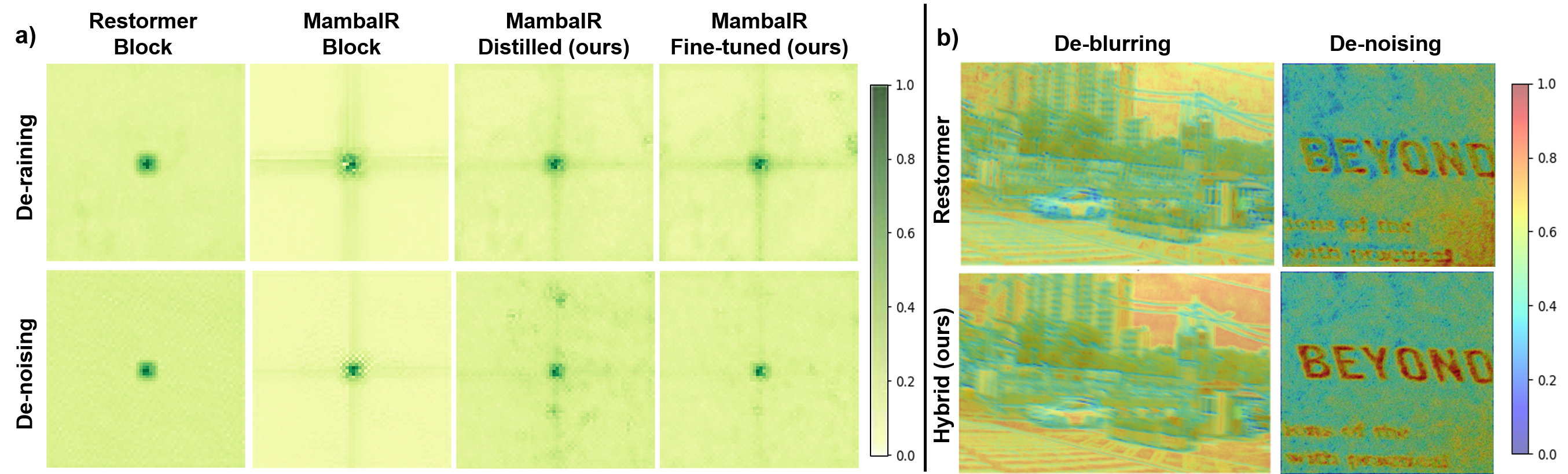}
\caption{(a) ERF visualization during hybrid training. Restormer shows global coverage, while standalone MambaIR exhibits limited spatial context. Feature distillation expands the MambaIR ERF, and after end-to-end fine-tuning it closely matches Restormer, indicating successful transfer of global context.
(b) Attention heatmaps of Restormer blocks. In the hybrid model, attention becomes more globally distributed than in the pure Restormer, showing that MambaIR enhances spatial awareness of transformer layers and highlighting the complementary interaction between the two blocks.}
\label{fig1}
\end{figure*}
In addition, attention heatmaps (in Figure ~\ref{fig1}) show that SSM blocks help transformer layers become more globally focused, suggesting bidirectional complementarity: transformers help SSMs generalize globally, while SSMs sharpen transformer attention. Together, this highlights the strength of our hybrid formulation.

Motivated by this possibility, in this work, we propose a modular, block-wise hybrid framework in which each transformer block is paired with an SSM block trained as a feature-distilled digital twin. Rather than redesigning operators or co-training hybrid layers, we formulate hybridization as a distillation problem. We introduce a method called \textit{Efficient Network Search} (ENS), which transforms deep hybrid design into a shallow block selection task by operating over pre-trained transformer and SSM components. ENS avoids training during the search and maintains computational efficiency.

We select Restormer as the expressive baseline because it represents the upper bound of accuracy among efficient Transformer models, owing to its linear attention and hierarchical design. MambaIR is chosen as the efficient counterpart, as its state-space recurrence delivers strong scalability and device-level latency benefits. Together, they form a natural expressive–efficient pairing for hybridization

We cast hybrid architecture design as a multi-objective discrete optimization problem, jointly maximizing restoration quality (e.g., PSNR, SSIM) and minimizing architectural complexity (e.g., transformer count). To solve this, we adopt Bayesian Optimization~\cite{snoek2012practical} with Expected Hypervolume Improvement (EHVI) as the acquisition strategy, suitable for expensive, black-box search in low dimensions. While BO itself is standard, its use over a distillation-aligned block pool enables practical and scalable hybrid exploration. From the resulting Pareto front, we apply knee-point analysis to select candidates with strong accuracy-efficiency trade-offs and fine-tune them end-to-end.

We evaluate our approach on motion and defocus deblurring, denoising, and deraining—using datasets such as GoPro, RealBlur, HIDE, SIDD, Rain13K and Test100. Our method consistently outperforms or matches strong baselines while reducing computation, validating the effectiveness of ENS across diverse degradation scenarios. Our key contributions are as follows:
\begin{itemize}
    \item We propose a modular hybrid framework that combines transformer and SSM blocks within a U-Net backbone, instantiated with Restormer and MambaIR.
    \item We introduce a digital twin distillation strategy that enables SSM blocks to replicate transformer outputs, enabling modular, training-free block substitution.
    \item We formulate an efficient hybrid architecture search method (ENS), using multi-objective Bayesian Optimization over pre-aligned blocks.
    \item We demonstrate competitive or improved performance across five restoration tasks compared to transformer- or SSM-only baselines.
    \item We provide empirical insights showing how SSM and transformer blocks mutually benefit each other through distillation and hybrid composition.
    \item Our framework is general and architecture-agnostic, enabling scalable hybridization for any compatible expressive–efficient block pair.
\end{itemize}

\begin{figure*}
\centering
\includegraphics[width=1.0\linewidth]{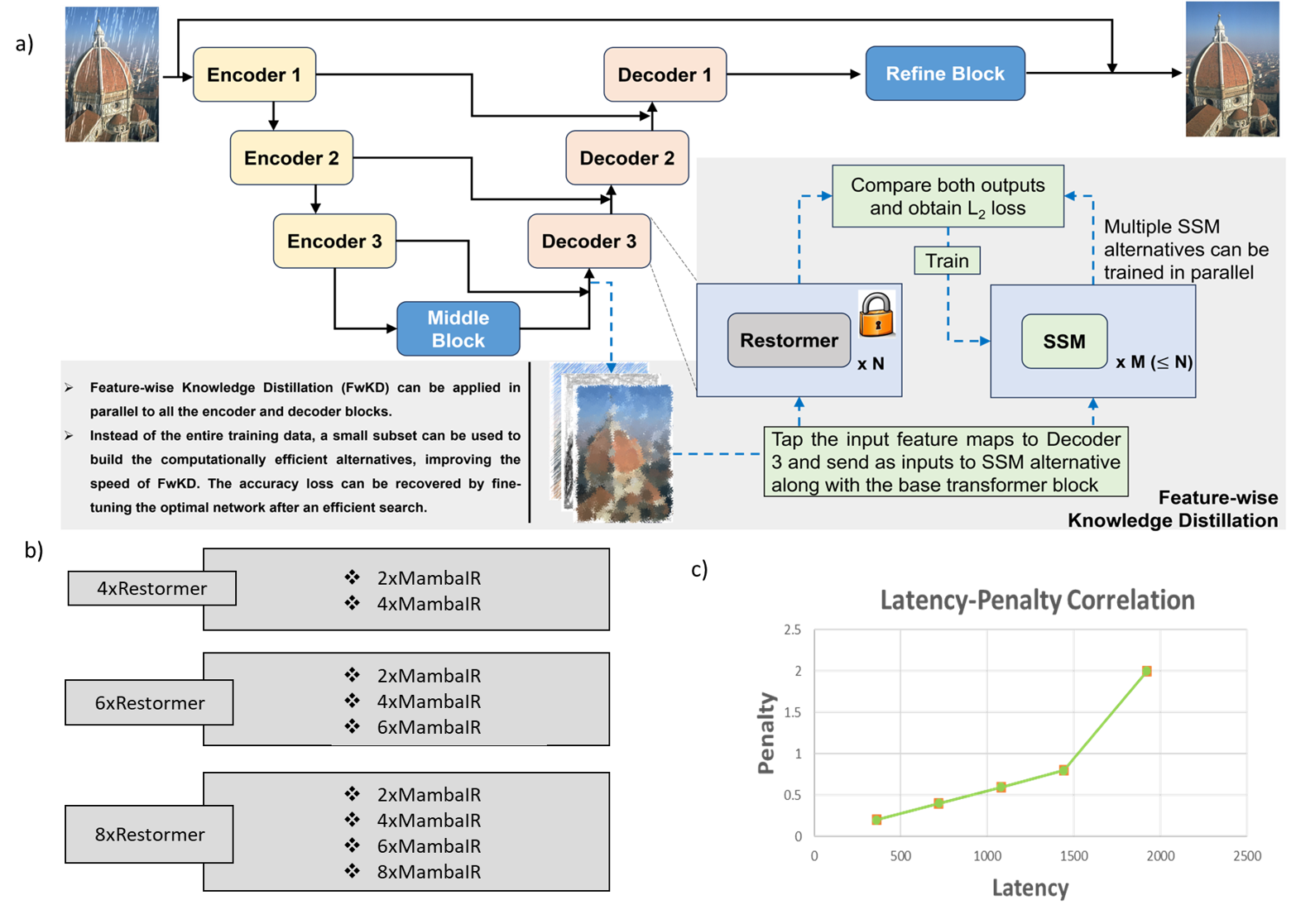}  
\caption{a) Pictorial representation of Feature-wise Knowledge Distillation b) Alternatives considered for each base block c) Correlation between latency and penalty }
\label{framework}
\end{figure*}

\section{Related Work}

Transformer-based architectures have transformed vision tasks by enabling long-range dependency modeling, beginning with ViT~\cite{dosovitskiy2020image} and extending to hierarchical variants like Swin Transformer~\cite{liu2021swin} and CSWin~\cite{dong2022cswin}. To reduce their quadratic complexity, several linear and efficient attention approximations have been proposed, including Linformer~\cite{wang2020linformer}, Performer~\cite{choromanski2020rethinking}, Longformer~\cite{beltagy2020longformer}, and BigBird~\cite{zaheer2020bigbird}. In the context of image restoration, task-specific transformer variants such as Restormer~\cite{zamir2022restormer}, HAT~\cite{chen2023hat}, and Uformer~\cite{wang2022uformer} adapt attention mechanisms to spatial hierarchies and encoder-decoder designs. Additional efforts like IPT~\cite{chen2021pre}, SwinIR~\cite{liang2021swinir}, DRSformer~\cite{lee2023drsformer}, and Rformer~\cite{luo2023rformer} improve transformer efficiency via multi-scale reasoning, cross-attention modules, or dual-branch fusion. Alongside these, lightweight convolutional baselines such as NAFNet~\cite{chen2022nafnet} and KBNet~\cite{zhang2023kbnet} are widely adopted for their simplicity and competitive performance in low-resource scenarios.

Recently, state-space models (SSMs), especially Mamba~\cite{gu2023mamba}, have gained traction for image restoration due to their linear-time inference and long-context modeling capacity. MambaIR~\cite{chen2024mambair} augments SSMs with local convolutions and channel attention to address spatial redundancy and pixel forgetting. MambaIRv2~\cite{mambairv2} introduces non-causal scanning for improved global context modeling. CU-Mamba~\cite{cu-mamba} employs spatial and channel-wise SSMs in a U-Net backbone for enhanced representational richness. MambaLLIE~\cite{mamballie} uses global-to-local SSMs for low-light enhancement, while TAMambaIR~\cite{tamambair} adds texture-aware modulation to handle localized degradations. RestorMamba~\cite{restormamba} addresses inpainting and denoising using Skip Scan and Enhanced Synergistic Mamba (ESM) blocks. Serpent~\cite{serpent} applies structured SSMs in multi-scale form for high-res restoration; Hi-Mamba~\cite{hi-mamba} reduces sequential bias with hierarchical direction-aware blocks. VmambaIR~\cite{vmambair} introduces Omni-Selective Scan, and DVMSR~\cite{dvmsr} distills transformer knowledge into compact Mamba-based models for super-resolution. TinyViM~\cite{tinyvim} incorporates frequency-aware routing to achieve resource-efficient designs. Surveys~\cite{survey1,survey2} consolidate these developments, framing Mamba as a strong alternative to attention-centric pipelines in restoration.

Hybrid models seek to combine the precision of transformers with the efficiency of SSMs. Jamba~\cite{jamba2024} integrates Mamba, transformer, and mixture-of-expert (MoE) layers to scale to long sequences. Samba~\cite{samba2024} alternates Mamba and sliding window attention for extrapolation over 1M tokens. Hymba~\cite{hymba2024} utilizes hybrid-heads combining attention and SSM within each layer, supplemented by meta tokens to improve memory and pattern recall. MambaVision~\cite{mambavision} fuses attention into SSM blocks for dense prediction. MatIR~\cite{li2024matir} and Contrast~\cite{contrast2024} alternate or interleave transformer and SSM modules at the block level, while TinyViM~\cite{tinyvim} routes low-frequency features through Mamba in compact dual-branch structures.

The rise of hybrid and SSM-only models illustrates growing interest in alternatives to attention-heavy architectures for image restoration. Most hybrid approaches fuse transformer and SSM functionality within a single block, inheriting both their strengths and limitations. In contrast, our work introduces a modular hybrid framework that decouples the two: transformer blocks are paired with feature-distilled SSM counterparts and selected via a training-free search. This hybridization strategy allows for better control over architectural composition and, to our knowledge, has not been previously explored in image restoration.

\section{Method}
\label{method}

We propose a modular hybrid framework for image restoration by integrating transformer and state-space model blocks within a unified U-Net backbone. Rather than designing new SSM variants or fusing attention and SSM into monolithic operators, we decouple the design into two tractable subproblems: feature-level distillation and block-wise architecture selection. Transformer blocks serve as expressive teachers, while Mamba-based SSM blocks are trained to mimic their behavior as lightweight digital twins. This enables a training-free, modular search for compact, task-adaptive architectures.

\subsection{Preliminaries}

In standard transformers, attention complexity scales quadratically with input resolution (H, W) as $\mathcal{O}(\hat{H}^2 \hat{W}^2)$, making them inefficient for high-resolution inputs. Restormer~\cite{zamir2022restormer} mitigates this by performing attention along channels, reducing spatial complexity to linear. Each Restormer block takes input $\mathbf{I} \in \mathbb{R}^{\hat{H} \times \hat{W} \times \hat{C}}$ and outputs $\mathbf{O} \in \mathbb{R}^{\hat{H} \times \hat{W} \times \hat{C}}$, using two key modules: Multi-DConv Head Transposed Attention (MDTA), which applies $1\times1$ convolutions, depth-wise spatial convolutions, and channel-wise attention; and Gated-DConv Feed-Forward Network (GDFN), which fuses two linear paths (one passed through GELU) via Hadamard product and applies a depth-wise convolution for spatial refinement. Both modules are wrapped with residual connections and LayerNorm~\cite{zamir2022restormer}.

MambaIR~\cite{chen2024mambair} blocks also map $\mathbf{I} \in \mathbb{R}^{\hat{H} \times \hat{W} \times \hat{C}}$ to $\mathbf{O} \in \mathbb{R}^{\hat{H} \times \hat{W} \times \hat{C}}$ via stacked Residual State Space Blocks (RSSBs), each comprising a Vision SSM (VSSM) with residual and normalization layers. VSSM flattens feature maps using four directional 1D scans (e.g., top-left to bottom-right), applies state-space modeling per scan, and aggregates the outputs. It includes two parallel branches: one processes features with depth-wise convolution and 2D SSM; the other uses a linear layer with SiLU activation. Outputs are fused via Hadamard product and passed through channel attention.

\subsection{Overall Architecture}

The backbone follows a four-level U-Net architecture. An input image $\mathbf{X} \in \mathbb{R}^{H \times W \times 3}$ is passed through a shallow convolution to obtain $\mathbf{F}_0 = \text{Conv}_{\text{stem}}(\mathbf{X})$. Three encoder stages $\mathcal{E}_1$, $\mathcal{E}_2$, $\mathcal{E}_3$ generate hierarchical features $\mathbf{F}_1$, $\mathbf{F}_2$, $\mathbf{F}_3$, followed by a bottleneck block $\mathcal{B}$ yielding $\mathbf{F}_4$. Between encoder stages, pixel-unshuffle halves spatial resolution and doubles channels.

The decoder upsamples features via pixel-shuffle, passes them through $\mathcal{D}_3$, $\mathcal{D}_2$, and $\mathcal{D}_1$, and fuses skip-connected encoder outputs using concatenation followed by $1\times1$ convolution. This produces $\mathbf{G}_1$, refined via a block $\mathcal{R}$ to yield $\hat{\mathbf{F}} = \mathcal{R}(\mathbf{G}_1)$. The restored image is given by $\hat{\mathbf{X}} = \mathbf{X} + \hat{\mathbf{F}}$. The model is optimized using an $\ell_1$ reconstruction loss: $\mathcal{L}_{\text{rec}} = | \hat{\mathbf{X}} - \mathbf{X}_{\text{GT}} |_1$. Following Restormer, we adopt a progressive learning strategy: training begins with small patches and gradually increases to full-resolution crops, adjusting batch size to maintain epoch time.

\subsection{Feature-wise Distillation}
\label{fwkd}

Each encoder, decoder, and bottleneck stage in the transformer teacher contains multiple identical Restormer blocks. To enable modular substitution and training-free design, we align each MambaIR block with its Restormer counterpart via feature-level knowledge distillation (see Figure \ref{framework} detailed pictorial description). Given input $\mathbf{I}$, we compute $\mathbf{O}_T = \text{Restormer}(\mathbf{I})$ and $\mathbf{O}_S = \text{MambaIR}(\mathbf{I})$. We minimize the loss $\mathcal{L}_{\text{distill}} = | \mathbf{O}_S - \mathbf{O}_T |_2^2$ to make MambaIR blocks serve as lightweight surrogates. As each pair is trained independently, this process is scalable and parallelizable, reducing the hybrid design problem to selection over pre-aligned modules.

\subsection{Hybrid Search Space Design}
\label{hwss}

We define an 8-dimensional search space based on the fixed Restormer configuration: [4, 6, 6, 8] in encoder and bottleneck, and [6, 6, 4, 4] in decoder and refinement. Instead of one-to-one replacements, we consider configurations where MambaIR blocks are used in full, partial, or compact forms as shown in Figure \ref{framework}. For example, a 6-block Restormer stage may be replaced by 6, 4, or 2 MambaIR blocks. This yields a search space of $3 \times 4 \times 4 \times 5 \times 4 \times 4 \times 3 \times 3 = 34{,}560$ hybrid configurations.

To support this, we pretrain 22 MambaIR surrogates: 2 for encoder-1, 3 each for encoder-2, encoder-3, decoder-3, decoder-2, 4 for the bottleneck, and 2 for decoder-1 and refinement. Each block is trained independently via distillation. This allows Efficient Network Search (ENS) to explore a large hybrid space using only a small, pre-aligned component pool. The obtained search space is tabulated in Supplementary file. 

\subsection{Efficient Network Search}

We formulate hybrid design as a discrete multi-objective optimization problem. Each architecture is encoded by an 8-dimensional vector $\mathbf{z} = [z_1, \ldots, z_8]$, where $z_i$ is a categorical variable denoting block type in stage $i$. The first objective is difference in PSNR$(\mathbf{z})$ and PSNR of base transformer model, measured on a fixed validation set. The second is Penalty$(\mathbf{z})$, which quantifies architectural complexity by summing penalties based on block type: full Restormers incur the highest, followed by large and small MambaIR variants as represented in Figure \ref{framework}. This yields the formulation:
\begin{equation}
\resizebox{\columnwidth}{!}{%
$\displaystyle \min_{\mathbf{z} \in \mathbb{Z}^8} \left( \text{PSNR}_{\text{Difference}}(\mathbf{z})\right) and \min_{\mathbf{z} \in \mathbb{Z}^8} \left(\text{Penalty}(\mathbf{z}) \right)
\quad \text{subject to} \quad \mathbf{z}_{\min} \leq \mathbf{z} \leq \mathbf{z}_{\max}$%
}
\label{eq:optimization}
\end{equation}

Although networks are composed via plug-and-play block integration, evaluating PSNR on high-resolution validation sets is computationally expensive. Both objectives are black-box and gradient-free, motivating the use of Bayesian Optimization (BO). We map the categorical space to a continuous vector $\mathbf{x} \in [0,1]^8$, which is discretized into $\mathbf{z}$ via fixed thresholds. BO models both objectives with Gaussian Processes and iteratively selects $\mathbf{x}$ by maximizing the Expected Hypervolume Improvement (EHVI)~\cite{snoek2012practical}. The resulting closed-loop search defines ENS: a training-free, sample-efficient strategy for discovering Pareto-optimal hybrid architectures.

To select the final architecture, we perform knee point analysis on the Pareto front to identify the configuration offering the best trade-off between PSNR and complexity. The selected model is then fine-tuned end-to-end, which helps recover accuracy losses arising from cascading mismatches introduced during surrogate-based block substitution in the search process.

\section{Experiments and Results}

\paragraph{Implementation Details.}
Our experiments are built upon the Restormer~\cite{zamir2022restormer} architecture, which serves as the teacher model for all hybrid designs. The fixed block configuration is [4, 6, 6, 8] in the encoder and bottleneck, and [6, 6, 4, 4] in the decoder and refinement stages, resulting in an 8-dimensional search space and $34{,}560$ possible configurations. To support this, we train 22 MambaIR surrogates in total—2 to 8 variants per stage, depending on block count. Each surrogate is trained independently via feature-level knowledge distillation, using fixed input–output pairs obtained from the pretrained Restormer model. This exercise required approximately 20 A100 hours. We adopt multi-objective Bayesian Optimization (BO) to explore the hybrid architecture space, modeling PSNR as the primary objective and block-wise penalty as the secondary objective. We start with $2*Dimensions + 1 = 17$ initial points distributed uniformly across the search space, and the BO is run for 500 function evaluations. From each resulting Pareto front, we select the five closest candidates near the knee and fine-tune each of them end-to-end following the same training schedule as the Restormer~\cite{zamir2022restormer} baseline. This includes progressive patch size scheduling, $\ell_1$ loss supervision, and validation on full-resolution images. The final model per task is chosen based on the best validation PSNR among the knee-region candidates.

\begin{table}[tb]
\centering
\caption{Comparison of Image deraining between different State-of-the-arts based on the considered baselines. The best results are highlighted in \textcolor{red}{red}, second‑best in \textcolor{blue}{blue}, and third‑best in \textcolor{green}{green}.}
\label{table1}
\resizebox{\textwidth}{!}{
\footnotesize 
\small
\begin{tabular}{@{}lcccccccccccc@{}} 
\toprule
\textbf{Method} &
\multicolumn{2}{c}{\textbf{Test100~\cite{li2019single}}} &
\multicolumn{2}{c}{\textbf{Rain100H~\cite{yang2017deep}}} &
\multicolumn{2}{c}{\textbf{Rain100L~\cite{yang2017deep}}} &
\multicolumn{2}{c}{\textbf{Test2800~\cite{jiang2020multi}}} &
\multicolumn{2}{c}{\textbf{Test1200~\cite{jiang2020multi}}} &
\multicolumn{2}{c}{\textbf{Average}} \\
\cmidrule(lr){2-3} \cmidrule(lr){4-5} \cmidrule(lr){6-7}
\cmidrule(lr){8-9} \cmidrule(lr){10-11} \cmidrule(lr){12-13}
 & PSNR\textsuperscript{$\uparrow$} & SSIM\textsuperscript{$\uparrow$} & PSNR\textsuperscript{$\uparrow$} & SSIM\textsuperscript{$\uparrow$} & PSNR\textsuperscript{$\uparrow$} & SSIM\textsuperscript{$\uparrow$} & PSNR\textsuperscript{$\uparrow$} & SSIM\textsuperscript{$\uparrow$} & PSNR\textsuperscript{$\uparrow$} & SSIM\textsuperscript{$\uparrow$} & PSNR\textsuperscript{$\uparrow$} & SSIM\textsuperscript{$\uparrow$} \\
\midrule
DerainNet~\cite{fu2017clearing}   & 22.77 & 0.810 & 14.92 & 0.592 & 27.03 & 0.884 & 24.31 & 0.861 & 23.38 & 0.835 & 22.48 & 0.796 \\
UMRLL~\cite{yasarla2019uncertainty}        & 24.41 & 0.829 & 26.01 & 0.832 & 29.18 & 0.923 & 29.97 & 0.905 & 30.55 & 0.910 & 28.02 & 0.880 \\
RESCAN~\cite{li2018recurrent}      & 25.00 & 0.835 & 26.36 & 0.786 & 29.80 & 0.881 & 31.29 & 0.904 & 30.51 & 0.882 & 28.59 & 0.857 \\
PreNet~\cite{ren2019progressive}      & 24.81 & 0.851 & 26.77 & 0.858 & 32.44 & 0.950 & 31.75 & 0.916 & 31.36 & 0.911 & 29.42 & 0.897 \\
MSPFN~\cite{jiang2020multi}       & 27.50 & 0.876 & 28.66 & 0.860 & 32.40 & 0.933 & 32.82 & 0.930 & 32.39 & 0.916 & 30.75 & 0.903 \\
SPAIR~\cite{purohit2021spatially}       & \textcolor{green}{30.35} & \textcolor{blue}{0.909} & 30.95 & 0.892 & 36.93 & 0.969 & 33.34 & 0.936 & 33.04 & 0.922 & 32.91 & 0.926 \\
MPRNET~\cite{zamir2021multi}       & 30.27 & \textcolor{green}{0.897} & 30.41 & 0.890 & 36.40 & 0.965 & 33.64 & \textcolor{green}{0.938} & 32.91 & 0.916 & 32.73 & 0.921 \\
Restormer~\cite{zamir2022restormer}   & \textcolor{blue}{32.00} & \textcolor{red}{0.923} & 31.46 & \textcolor{green}{0.904} & 38.99 & 0.978 & \textcolor{blue}{34.18} & \textcolor{blue}{0.944} & \textcolor{green}{33.19} & \textcolor{blue}{0.926} & \textcolor{blue}{33.96} & \textcolor{blue}{0.935} \\
Fourmer~\cite{zhou2023fourmer}     & NA & NA & 30.76 & 0.896 & 37.47 & 0.970 & NA & NA & 33.05 & 0.921 & NA & NA \\
IR‑SDE~\cite{luo2023image}      & NA & NA & 31.65 & 0.904 & 38.30 & \textcolor{blue}{0.980} & 30.42 & 0.891 & NA & NA & NA & NA \\
MambaIR~\cite{chen2024mambair}     & NA & NA & 30.62 & 0.893 & 38.78 & 0.977 & 33.58 & 0.927 & 32.56 & \textcolor{green}{0.923} & \textcolor{green}{33.89} & \textcolor{green}{0.930} \\
VMambaIR~\cite{vmambair}    & NA & NA & \textcolor{green}{31.66} & \textcolor{blue}{0.909} & \textcolor{green}{39.09} & \textcolor{green}{0.979} & 34.01 & \textcolor{blue}{0.944} & \textcolor{blue}{33.33} & \textcolor{blue}{0.926} & NA & NA \\
FreqMamba~\cite{zou2024freqmamba}   & NA & NA & \textcolor{blue}{31.74} & \textcolor{red}{0.912} & \textcolor{blue}{39.18} & \textcolor{red}{0.981} & \textcolor{red}{34.25} & \textcolor{red}{0.951} & \textcolor{red}{33.36} & \textcolor{red}{0.931} & NA & NA \\
\midrule
\textbf{Ours} &
\textcolor{red}{32.03} & \textcolor{red}{0.923} &
\textcolor{red}{31.86} & \textcolor{red}{0.912} &
\textcolor{red}{39.37} & \textcolor{green}{0.979} &
\textcolor{green}{34.13} & \textcolor{blue}{0.944} &
32.85 & \textcolor{green}{0.923} &
\textcolor{red}{34.05} & \textcolor{red}{0.936} \\
\bottomrule
\end{tabular}
} 
\end{table}

\subsection{Edge Efficiency}

A central motivation of our work is that FLOPs and GPU memory—commonly used as efficiency proxies—do not reliably predict latency on edge hardware. This mismatch is particularly pronounced for attention-based models, where operations that appear efficient in theory can incur substantial runtime overhead on mobile SoCs. Although Restormer is highly FLOP- and memory-efficient due to linear attention (Fig.~\ref{fig0}), its attention kernels lead to significant latency during on-device execution. To address this, our ENS objective incorporates a lightweight transformer-usage penalty as a surrogate for latency, enabling fully training-free architecture search without repeatedly profiling thousands of candidates on hardware.

The effectiveness of this design is evident on the Snapdragon~8~Elite CPU. Restormer requires \textbf{10119.52 ms} for inference, whereas the ENS-discovered hybrids achieve substantially lower latency. \textbf{ENS-Deblurring} reduces runtime to \textbf{2973 ms}, delivering a \textbf{3.4$\times$ speedup} over Restormer. \textbf{ENS-Deraining} achieves \textbf{5816 ms}, corresponding to a \textbf{1.74$\times$ improvement}, while \textbf{ENS-Denoising} runs in \textbf{8666 ms}, yielding a \textbf{1.17$\times$ improvement}. Importantly, these latency reductions are achieved while maintaining competitive restoration quality.

These results highlight that FLOPs and memory alone do not reflect real runtime behavior on edge devices, and demonstrate that the penalty-based ENS strategy effectively discovers hybrid architectures that provide meaningful practical efficiency improvements over pure transformer designs.

\subsection{Deraining Performance}

We evaluate our framework on single-image deraining using the Rain13K dataset for training, and test it across five diverse benchmarks: Rain100H, Rain100L, Test100, Test1200, and Test2800. We compare the performance of the Restormer baseline, MambaIR-only model, and the hybrid architecture discovered through ENS in Table ~\ref{table1} .

Our hybrid model consistently outperforms the MambaIR-only baseline and performs competitively with Restormer, despite using fewer transformer blocks. The improvement is particularly noticeable on complex scenes with dense or fine-grained rain streaks as shown in Figure ~\ref{fig2}. The hybrid architecture benefits from Restormer's contextual richness and Mamba's memory efficiency, leading to more balanced performance across datasets. Visual comparisons further reveal that our model retains more background texture while effectively removing rain artifacts.

\subsection{Deblurring Performance}

We evaluate the hybrid framework on two types of deblurring: motion and defocus. For motion deblurring, the model is trained on the GoPro dataset and tested on RealBlur and HIDE, representing cross-domain generalization. For defocus deblurring, the architecture selected from the ENS process is fine-tuned on the DPDD~\cite{abuolaim2020defocus} dataset.

\begin{table*}[htbp]
\centering
\caption{Comparison of Motion deblurring between different State-of-the-arts based on the considered baselines. The best results are highlighted in \textcolor{red}{red}, second‑best in \textcolor{blue}{blue}, and third‑best in \textcolor{green}{green}.}
\label{table2}
\footnotesize
\begin{tabular}{@{}lcccccccc@{}} 
\toprule
\textbf{Method} & 
\multicolumn{2}{c}{\textbf{GoPro~\cite{nah2017deep}}} &
\multicolumn{2}{c}{\textbf{HIDE~\cite{shen2019human}}} &
\multicolumn{2}{c}{\textbf{RealBlur-J~\cite{rim2020real}}} &
\multicolumn{2}{c}{\textbf{RealBlur-R~\cite{rim2020real}}} \\
\cmidrule(lr){2-3} \cmidrule(lr){4-5} \cmidrule(lr){6-7} \cmidrule(lr){8-9}
 & PSNR\textsuperscript{$\uparrow$} & SSIM\textsuperscript{$\uparrow$} & PSNR\textsuperscript{$\uparrow$} & SSIM\textsuperscript{$\uparrow$} & PSNR\textsuperscript{$\uparrow$} & SSIM\textsuperscript{$\uparrow$} & PSNR\textsuperscript{$\uparrow$} & SSIM\textsuperscript{$\uparrow$} \\
\midrule
Xu et al.~\cite{xu2013unnatural} & 21.00 & 0.741 & NA & NA & 27.14 & 0.830 & 34.46 & 0.937 \\
DeblurGAN~\cite{kupyn2018deblurgan}     & 28.70 & 0.858 & 24.51 & 0.871 & 27.97 & 0.834 & 33.79 & 0.903 \\
Nah et al.~\cite{nah2017deep}. & 29.08 & 0.914 & 25.73 & 0.874 & 27.87 & 0.827 & 32.51 & 0.841 \\
Zhang et al.~\cite{zhang2018dynamic}. & 29.19 & 0.931 & NA & NA & 27.80 & 0.847 & 35.48 & 0.947 \\
DeblurGAN-v2~\cite{kupyn2019deblurgan} & 29.55 & 0.934 & 26.61 & 0.875 & 28.70 & 0.866 & 35.26 & 0.944 \\
SRN~\cite{tao2018scale}  & 30.26 & 0.934 & 28.36 & 0.915 & 28.56 & 0.867 & 35.66 & 0.947 \\
DBGAN~\cite{zhang2020deblurring} & 31.10 & 0.942 & 28.94 & 0.915 & 24.93 & 0.745 & 33.78 & 0.909 \\
MT-RNN~\cite{park2020multi} & 31.15 & 0.945 & 29.15 & 0.918 & 28.44 & 0.862 & 35.79 & 0.951 \\
DMPHN~\cite{zhang2019deep} & 31.20 & 0.940 & 29.09 & 0.924 & 28.42 & 0.860 & 35.70 & 0.948 \\
SPAIR~\cite{purohit2021spatially}  & 32.06 & 0.953 & 30.29 & 0.931 & \textcolor{green}{28.81} & \textcolor{green}{0.875} & NA & NA \\
MIMO-Unet+~\cite{cho2021rethinking} & 32.45 & 0.957 & 29.99 & 0.930 & 27.63 & 0.837 & 35.54 & 0.947 \\
MPRNet~\cite{zamir2021multi} & \textcolor{green}{32.66} & \textcolor{green}{0.959} & \textcolor{green}{30.96} & \textcolor{green}{0.939} & 28.70 & 0.873 & 35.99 & 0.952 \\
Restormer~\cite{zamir2022restormer}  & \textcolor{blue}{32.92} & \textcolor{blue}{0.961} & \textcolor{blue}{31.22} & \textcolor{blue}{0.942} & \textcolor{blue}{28.96} & \textcolor{blue}{0.879} & \textcolor{blue}{36.19} & \textcolor{blue}{0.957} \\
MambaIR(ours) & 31.56 & 0.950 & 30.29 & 0.909 & 28.76 & 0.871 & \textcolor{green}{36.00} & \textcolor{green}{0.953} \\
\midrule
\textbf{Ours} & \textcolor{red}{33.08} & \textcolor{red}{0.962} & \textcolor{red}{31.25} & \textcolor{red}{0.943} & \textcolor{red}{30.22} & \textcolor{red}{0.901} & \textcolor{red}{36.76} & \textcolor{red}{0.959} \\
\bottomrule
\end{tabular}
\end{table*}

This is the first known application of a Mamba-based architecture to image deblurring (mentioned as ours in Table ~\ref{table2}. Our hybrid approach achieves comparable or superior PSNR and SSIM scores relative to the Restormer baseline, with improved performance on unseen test sets like RealBlur (see Table ~\ref{table2} and Table ~\ref{table3}). The hybrid architecture demonstrates strong generalization, aided by the MambaIR blocks' efficient temporal memory and the transformer blocks’ ability to capture global image structure. As presented in Figure ~\ref{fig2} the output images show sharper edges and fewer residual blur artifacts.

\begin{table*}[htbp]
\centering
\caption{Comparison of Defocus deblurring between different State-of-the-arts based on the considered baselines, where subscripts S and D stand for single and dual-pixel defocus deblurring, respectively. Best results are in \textcolor{red}{red}, second-best in \textcolor{blue}{blue}, and third-best in \textcolor{green}{green}.}
\label{table3}
\resizebox{\textwidth}{!}{
\scriptsize
\begin{tabular}{@{}lcccccccccccc@{}} 
\toprule
\textbf{Method} & 
\multicolumn{4}{c}{\textbf{Outdoor}} & 
\multicolumn{4}{c}{\textbf{Indoor}} & 
\multicolumn{4}{c}{\textbf{Combined}} \\
\cmidrule(lr){2-5} \cmidrule(lr){6-9} \cmidrule(lr){10-13}
 & PSNR\textsuperscript{$\uparrow$} & SSIM\textsuperscript{$\uparrow$} & MAE\textsuperscript{$\downarrow$} & LPIPS\textsuperscript{$\downarrow$} & PSNR\textsuperscript{$\uparrow$} & SSIM\textsuperscript{$\uparrow$} & MAE\textsuperscript{$\downarrow$} & LPIPS\textsuperscript{$\downarrow$} & PSNR\textsuperscript{$\uparrow$} & SSIM\textsuperscript{$\uparrow$} & MAE\textsuperscript{$\downarrow$} & LPIPS\textsuperscript{$\downarrow$} \\
\midrule
\multicolumn{13}{l}{\textit{Single-pixel (S) Methods}} \\
JNB\textsubscript{S}~\cite{shi2015just} & 21.10 & 0.608 & 0.064 & 0.355 & 26.73 & 0.828 & 0.031 & 0.273 & 23.84 & 0.715 & 0.048 & 0.315 \\
EBDB\textsubscript{S}~\cite{karaali2017edge} & 21.25 & 0.599 & 0.058 & 0.373 & 25.77 & 0.772 & 0.040 & 0.297 & 23.45 & 0.683 & 0.049 & 0.336 \\
DMENet\textsubscript{S}~\cite{lee2019deep} & 21.43 & 0.644 & 0.063 & 0.397 & 25.50 & 0.788 & 0.038 & 0.298 & 23.41 & 0.714 & 0.051 & 0.349 \\
DPDNet\textsubscript{S}~\cite{abuolaim2020defocus} & 22.25 & 0.682 & 0.056 & 0.313 & 26.54 & 0.816 & 0.031 & 0.239 & 24.43 & 0.747 & 0.044 & 0.277 \\
KPAC\textsubscript{S}~\cite{son2021single} & 22.62 & 0.701 & 0.053 & 0.269 & 27.97 & 0.852 & \textcolor{green}{0.026} & 0.182 & 25.22 & 0.774 & 0.040 & 0.227 \\
IFAN\textsubscript{S}~\cite{lee2021iterative} & 22.76 & 0.720 & \textcolor{green}{0.052} & 0.254 & 28.11 & 0.861 & \textcolor{green}{0.026} & 0.179 & 25.37 & 0.789 & \textcolor{green}{0.039} & 0.217 \\
KBNet\textsubscript{S}~\cite{zhang2023kbnet} & \textcolor{green}{23.10} & 0.736 & \textcolor{red}{0.050} & \textcolor{blue}{0.233} & \textcolor{blue}{28.87} & \textcolor{red}{0.882} & \textcolor{blue}{0.025} & \textcolor{red}{0.143} & NA & NA & NA & NA \\
Restormer\textsubscript{S}~\cite{zamir2022restormer} & \textcolor{blue}{23.24} & \textcolor{red}{0.743} & \textcolor{red}{0.050} & \textcolor{red}{0.209} & \textcolor{blue}{28.87} & \textcolor{red}{0.882} & \textcolor{blue}{0.025} & \textcolor{blue}{0.145} & \textcolor{blue}{25.98} & \textcolor{red}{0.811} & \textcolor{blue}{0.038} & \textcolor{red}{0.178} \\
MambaIR\textsubscript{S}(ours) & 22.99 & \textcolor{green}{0.739} & \textcolor{blue}{0.051} & 0.278 & \textcolor{green}{28.25} & \textcolor{green}{0.874} & \textcolor{red}{0.024} & 0.190 & \textcolor{green}{25.55} & \textcolor{green}{0.805} & \textcolor{blue}{0.038} & \textcolor{green}{0.235} \\
\textbf{Ours}\textsubscript{S} & \textcolor{red}{23.34} & \textcolor{blue}{0.741} & \textcolor{red}{0.050} & \textcolor{green}{0.243} & \textcolor{red}{29.14} & \textcolor{blue}{0.876} & \textcolor{red}{0.024} & \textcolor{green}{0.174} & \textcolor{red}{26.16} & \textcolor{blue}{0.807} & \textcolor{red}{0.037} & \textcolor{blue}{0.209} \\
\midrule
\multicolumn{13}{l}{\textit{Dual-pixel (D) Methods}} \\
RDPD\textsubscript{D}~\cite{abuolaim2021learning} & 22.82 & 0.704 & 0.053 & 0.298 & 28.10 & 0.843 & 0.027 & 0.210 & 25.39 & 0.772 & 0.040 & 0.255 \\
DPDNet\textsubscript{D}~\cite{abuolaim2020defocus} & 22.90 & 0.726 & 0.052 & 0.255 & 27.48 & 0.849 & 0.029 & 0.189 & 25.13 & 0.786 & 0.041 & 0.223 \\
Uformer\textsubscript{D}~\cite{wang2022uformer} & 23.10 & 0.728 & 0.051 & 0.285 & 28.23 & 0.860 & \textcolor{green}{0.026} & 0.199 & 25.65 & 0.795 & \textcolor{green}{0.039} & 0.243 \\
IFAN\textsubscript{D}~\cite{lee2021iterative} & \textcolor{green}{23.46} & \textcolor{green}{0.743} & \textcolor{green}{0.049} & \textcolor{green}{0.240} & \textcolor{green}{28.66} & \textcolor{green}{0.868} & \textcolor{blue}{0.025} & \textcolor{green}{0.172} & \textcolor{green}{25.99} & \textcolor{green}{0.804} & \textcolor{blue}{0.037} & \textcolor{green}{0.207} \\
Restormer\textsubscript{D}~\cite{zamir2022restormer} & \textcolor{blue}{23.97} & \textcolor{red}{0.773} & \textcolor{blue}{0.047} & \textcolor{red}{0.175} & \textcolor{blue}{29.48} & \textcolor{red}{0.895} & \textcolor{red}{0.023} & \textcolor{red}{0.134} & \textcolor{blue}{26.66} & \textcolor{red}{0.833} & \textcolor{red}{0.035} & \textcolor{red}{0.155} \\
MambaIR\textsubscript{D}(ours) & 22.39 & 0.699 & 0.053 & 0.342 & 27.05 & 0.845 & 0.028 & 0.240 & 24.66 & 0.770 & 0.041 & 0.292 \\
\textbf{Ours}\textsubscript{D} & \textcolor{red}{24.00} & \textcolor{blue}{0.768} & \textcolor{red}{0.046} & \textcolor{blue}{0.200} & \textcolor{red}{29.63} & \textcolor{blue}{0.893} & \textcolor{red}{0.023} & \textcolor{blue}{0.147} & \textcolor{red}{26.74} & \textcolor{blue}{0.829} & \textcolor{red}{0.035} & \textcolor{blue}{0.174} \\
\bottomrule
\end{tabular}
}
\end{table*}

\begin{table*}[]
\caption{Comparison of Real Denoising between different State-of-the-arts based on the considered baselines. Best results are in \textcolor{red}{red}, second-best in \textcolor{blue}{blue}, and third-best in \textcolor{green}{green}.}
\label{table4}
\centering
\small
\resizebox{\textwidth}{!}{%
\begin{tabular}{ccccccccccccccccccccc}
\toprule
{Dataset} &
  \multicolumn{2}{c}{DeamNet~\cite{ren2021adaptive}} &
  \multicolumn{2}{c}{MIRNet~\cite{zamir2020learning}} &
  \multicolumn{2}{c}{MPRNet~\cite{zamir2021multi}} &
  \multicolumn{2}{c}{NBNET~\cite{cheng2021nbnet}} &
  \multicolumn{2}{c}{Uformer~\cite{wang2022uformer}} &
  \multicolumn{2}{c}{Restormer~\cite{zamir2022restormer}} &
  \multicolumn{2}{c}{MambaIR~\cite{chen2024mambair}} &
  \multicolumn{2}{c}{MaIR~\cite{li2024mair}} &
  \multicolumn{2}{c}{Ours} \\
  \cmidrule(lr){2-3} \cmidrule(lr){4-5} \cmidrule(lr){6-7} \cmidrule(lr){8-9} \cmidrule(lr){10-11} \cmidrule(lr){12-13} \cmidrule(lr){14-15} \cmidrule(lr){16-17} \cmidrule(lr){18-19}
 &
  PSNR\textsuperscript{$\uparrow$} &
  SSIM\textsuperscript{$\uparrow$} &
  PSNR\textsuperscript{$\uparrow$} &
  SSIM\textsuperscript{$\uparrow$} &
  PSNR\textsuperscript{$\uparrow$} &
  SSIM\textsuperscript{$\uparrow$} &
  PSNR\textsuperscript{$\uparrow$} &
  SSIM\textsuperscript{$\uparrow$} &
  PSNR\textsuperscript{$\uparrow$} &
  SSIM\textsuperscript{$\uparrow$} &
  PSNR\textsuperscript{$\uparrow$} &
  SSIM\textsuperscript{$\uparrow$} &
  PSNR\textsuperscript{$\uparrow$} &
  SSIM\textsuperscript{$\uparrow$} &
  PSNR\textsuperscript{$\uparrow$} &
  SSIM\textsuperscript{$\uparrow$} &
  PSNR\textsuperscript{$\uparrow$} &
  SSIM\textsuperscript{$\uparrow$} & \\
\midrule
\textbf{SIDD}~\cite{abdelhamed2018high} &
  39.47 &
  0.957 &
  39.72 &
  0.959 &
  39.71 &
  0.958 &
  39.75 &
  \textcolor{green}{0.959} &
  39.77 &
  \textcolor{green}{0.959} &
  \textcolor{blue}{40.02} &
  \textcolor{blue}{0.960} &
  39.89 &
  \textcolor{blue}{0.960} &
  \textcolor{green}{39.92} &
  \textcolor{blue}{0.960} &
  \textcolor{red}{40.04} &
  \textcolor{red}{0.961} \\
\midrule
\textbf{SenseNoise}~\cite{zhang2022idr} &
  NA &
  NA &
  39.3 &
  0.919 &
  \textcolor{green}{35.43} &
  \textcolor{green}{0.922} &
  NA &
  NA &
  35.43 &
  0.920 &
  \textcolor{blue}{35.52} &
  \textcolor{red}{0.924} &
  NA &
  NA &
  NA &
  NA &
  \textcolor{red}{35.87} &
  \textcolor{blue}{0.923} \\
\bottomrule
\end{tabular}%
}

\end{table*}

\subsection{Denoising Performance}

We evaluate our method on real image denoising, where the model is trained on SIDD and evaluated on SenseNoise. The hybrid architecture is obtained via ENS, and fine-tuned following the same progressive strategy used in Restormer. 

The hybrid model surpasses both MambaIR and Restormer baselines on real-noise benchmarks (see Table ~\ref{table4}), offering improved noise suppression while preserving fine details. The transformer blocks provide accurate global context modeling, whereas the MambaIR blocks help reduce redundancy and noise artifacts with lower overhead. In challenging conditions with strong degradations, our model produces cleaner and sharper outputs (see Figure ~\ref{fig2}).

\begin{figure*}
\centering
\includegraphics[width=\linewidth]{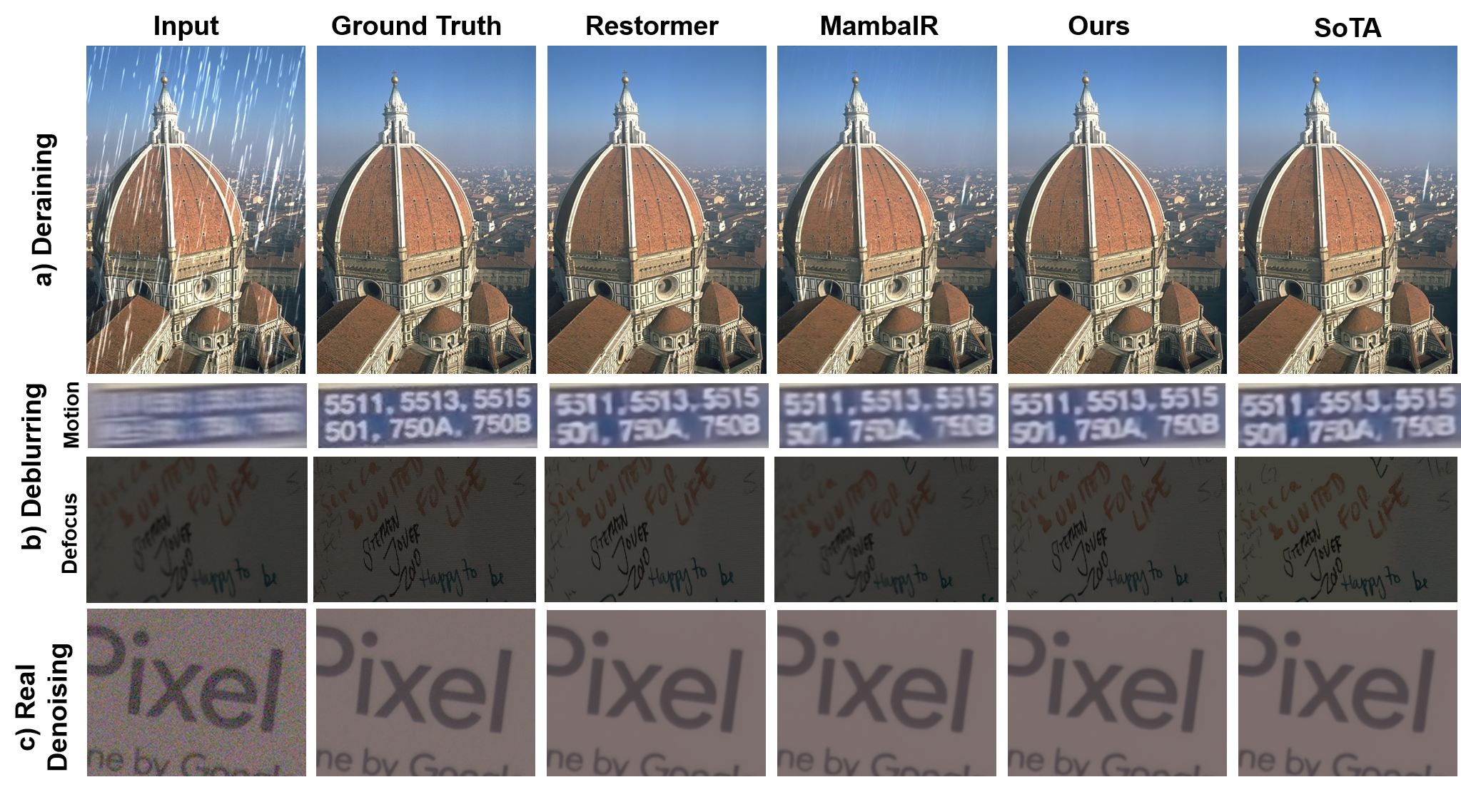}  
\caption{Visual Quality comparison across different Image Restoration tasks. State-of-the-art (SOTA) being compared in case of Deraining is MPRNet~\cite{zamir2021multi}, Motion Deblurring is DBGAN~\cite{zhang2020deblurring}, Defocus Deblurring is KBNet~\cite{zhang2023kbnet}, and Denoising is MaIR~\cite{li2024mair}}
\label{fig2}
\end{figure*}

\subsection{Ablation Analysis}
\paragraph{Impact of Feature-wise Distillation} Feature-wise knowledge distillation (FwKD) is central to our hybrid design framework, enabling the transformation of MambaIR blocks into lightweight surrogates of transformer blocks without full-model retraining. To evaluate its contribution, we perform an ablation where the ENS-optimized architectures are initialized with random weights and trained with progressive learning.

We observe a significant drop in validation PSNR across all tasks when FwKD is removed (see Table ~\ref{table5}), indicating that the surrogate blocks fail to capture the functional behavior of their transformer counterparts. This not only degrades individual block quality but also compromises the architecture search, as block-level substitutions no longer preserve performance under hybrid composition. Furthermore, attention heatmaps and receptive field analysis as presented in Figure ~\ref{fig1} shows narrower activation spread and weaker context modeling when FwKD is omitted. These results confirm that distillation is critical for convergence to desired accuracy and not only for enabling modular, training-free hybrid design, thus justifying its role in ENS.

\begin{table*}[htbp]
\centering
\caption{Ablation studies analyzing the effect of feature-wise knowledge distillation, impact of ENS, and justification of hybrid architecture.}
\label{table5}
\footnotesize
\resizebox{\textwidth}{!}{
\begin{tabular}{@{}lcccccccc@{}}
\toprule
\textbf{Ablation Analysis} &
\multicolumn{2}{c}{\textbf{GoPro~\cite{nah2017deep} - Deblur}} &
\multicolumn{2}{c}{\textbf{SIDD~\cite{abdelhamed2018high} - Denoise}} &
\multicolumn{2}{c}{\textbf{Rain100L~\cite{yang2017deep} - Derain}} \\
\cmidrule(r){2-3} \cmidrule(r){4-5} \cmidrule(r){6-7} \cmidrule(r){8-9}
\textbf{Design \& Training Configuration} & \textbf{PSNR}\textsuperscript{$\uparrow$} & \textbf{SSIM}\textsuperscript{$\uparrow$} & \textbf{PSNR}\textsuperscript{$\uparrow$} & \textbf{SSIM}\textsuperscript{$\uparrow$} & \textbf{PSNR}\textsuperscript{$\uparrow$} & \textbf{SSIM}\textsuperscript{$\uparrow$} \\
\midrule
\textit{\textbf{Impact of Feature-wise Knowledge Distillation}} \\
\quad Proposed model with FwKD & \textcolor{red}{33.08} & \textcolor{red}{0.962} & \textcolor{red}{40.04} & \textcolor{red}{0.961} & \textcolor{red}{39.37} & \textcolor{red}{0.979} \\
\quad Proposed model without FwKD & \textcolor{blue}{28.54} & \textcolor{blue}{0.913} & \textcolor{blue}{39.69} & \textcolor{blue}{0.958} & \textcolor{blue}{35.39} & \textcolor{blue}{0.960} \\[5pt] 
\midrule
\textit{\textbf{Impact of Efficient Network Search (ENS)}} \\
\quad Randomly designed Hybrid network & \textcolor{red}{31.82} & \textcolor{red}{0.952} & \textcolor{red}{39.88} & \textcolor{red}{0.960} & \textcolor{red}{39.98} & \textcolor{red}{0.977} \\
\quad Hybrid design with Equal blocks & \textcolor{blue}{31.02} & \textcolor{blue}{0.944} & \textcolor{blue}{39.85} & \textcolor{blue}{0.960} & \textcolor{blue}{37.51} & \textcolor{blue}{0.968} \\[5pt] 

\bottomrule
\end{tabular}
}
\end{table*}

\paragraph{Importance of Efficient Network Search}

To assess the effectiveness of our Efficient Network Search (ENS), we compare it against two alternative hybrid design strategies: (i) random selection of hybrid configurations from the same search space, and (ii) rule-based hybrids constructed by assigning 50\% of the blocks in each stage to Restormer and the remaining 50\% to MambaIR, similar to strategies used in prior hybrid vision models ~\cite{liu2025srmamba}. All configurations are composed using the same feature-distilled block pool to ensure fairness and are fine-tuned identically.

Hybrid architectures discovered via ENS consistently outperform both baselines in terms of PSNR and efficiency (see Table ~\ref{table5}), demonstrating better trade-offs between transformer usage and restoration quality. Random and rule-based designs tend to fall outside the knee region of the Pareto front (see the diagrams in the Supplementary file), often overusing transformer layers or failing to allocate sufficient expressivity to key stages. In contrast, ENS identifies architectures concentrated around the Pareto knee, where performance plateaus despite increasing complexity. 

\paragraph{Extending to another expressive-efficient - Uformer-MambaIR:}

\begin{table}
\centering
\caption{
Edge-efficiency of Uformer-SSM hybrids designed by ENS for denoising use-case. Once again, it can be seen that the proposed method generates architectures which balances the accuracy and latency (profiled on CPU of Qualcomm's SM8750 chipset on Samsung Galaxy S25 ultra in INT4 precision). We reiterate that, these architecture reach the accuracy of the transformer base models but have significant lower latencies on device balancing the trade-off efficiently. 
}
\label{uformer}
\footnotesize
\begin{tabular}{cccc}
\hline
\textbf{Model} & \textbf{PSNR} & \textbf{SSIM} & \textbf{Latency (ms)}        \\ \hline
Uformer        & 39.8          & 0.959         & 660                          \\ 
Arch-1         & 39.6          & 0.9586        & {\color[HTML]{FF0000} 437.8} \\ 
Arch-2         & 39.8          & 0.9589        & {\color[HTML]{FF0000} 489.8} \\ 
 \hline
\end{tabular}
\end{table}

To evaluate the generality of the proposed framework beyond the Restormer-MambaIR pairing, an ablation study was conducted using Uformer as the expressive backbone and MambaIR as the efficient surrogate. Results show that Uformer achieves high denoising accuracy but incurs significant latency (660 ms) on a Qualcomm SM8750 platform, while MambaIR offers lower latency (111 ms) but reduced PSNR. Applying the hardware-aware ENS formulation to this pair yielded hybrid architectures with near-identical accuracy to Uformer but reduced latency (437 ms and 489 ms). These hybrids balance expressive power and efficiency by leveraging Uformer in high-level stages and MambaIR in latency-critical stages, without architectural modifications to either model.

The study demonstrates that ENS is not specific to Restormer but generalizes robustly to other expressive models with different attention layouts and U-Net structures. The hardware-aware search space and feature-wise distillation (FwKD) remain effective across diverse backbones, enabling meaningful latency-accuracy trade-offs. These findings highlight the architecture-agnostic nature of the framework, driven by principled surrogate alignment and hardware-aware search, offering a systematic approach to edge-efficient image restoration.

\paragraph{Limitations}
Our framework presents a scalable path for efficient hybrid architectures. Key limitations include the assumption of sufficient block-wise functional substitution (without modeling deeper dependencies), the use of an indirect complexity proxy rather than direct latency measurement during search, and the inherent trade-off of training-free ENS between scalability and co-adaptation. These points suggest clear future research directions.

\section{Conclusion}

We presented a modular framework for hybrid architecture design that combines expressive transformer blocks with efficient state-space models for high-resolution image restoration. By framing the problem as a block-level distillation and selection task, we enabled plug-and-play substitution of transformer blocks with distilled MambaIR surrogates, converting deep sequence modeling into a shallow architectural optimization problem. Our Efficient Network Search (ENS) procedure leverages multi-objective Bayesian optimization over this compact space to discover task-adaptive hybrids without retraining during search. Through experiments on five restoration tasks, we demonstrated that our approach achieves strong performance with reduced computational cost, outperforming both transformer-only and SSM-only baselines. Analyses of receptive fields and attention maps reveal how distilled blocks inherit contextual capacity, while hybridization enables mutual enhancement between components. Our framework is architecture-agnostic and extensible to other expressive–efficient block pairings, offering a scalable path toward efficient and adaptable model design.

%
%
\bibliographystyle{splncs04}
\bibliography{main}

\end{document}